\documentclass[conference]{IEEEtran}
\IEEEoverridecommandlockouts

\usepackage{cite}
\usepackage{amsmath,amssymb,amsfonts}
\usepackage{algorithmic}
\usepackage{graphicx}
\usepackage{textcomp}
\usepackage{xcolor}
\usepackage{colortbl}
\usepackage{caption}   
\usepackage{hyperref} 
\usepackage{placeins}
\usepackage{subfigure} 
\usepackage{float}
\usepackage{multirow}

\def\BibTeX{{\rm B\kern-.05em{\sc i\kern-.025em b}\kern-.08em
    T\kern-.1667em\lower.7ex\hbox{E}\kern-.125emX}}
\begin{document}

\title{Frozen LLMs as Map-Aware Spatio-Temporal Reasoners for Vehicle Trajectory Prediction \\
}

\author{Yanjiao Liu, Jiawei Liu, Xun Gong, Zifei Nie*,
\thanks{This work is supported by National Nature Science Foundation of China under Grant (62573209), Jilin Provincial Science and Technology Development Project (20260601023RC)}
\thanks{All the authors are with School of Artificial Intelligence, Jilin University, Changchun 130012, PR China.}%
\thanks{*The Corresponding author is Zifei Nie{\tt\small (zifei\_nie@jlu.edu.cn)}}
}



\maketitle
\begin{abstract}
Large language models (LLMs) have recently demonstrated strong reasoning capabilities and attracted increasing research attention in the field of autonomous driving (AD). However, safe application of LLMs on AD's perception and prediction still requires a thorough understanding of both the  dynamic traffic agents and the static road infrastructure. To this end, this study introduces a framework to evaluate the capability of LLMs in understanding the behaviors of dynamic traffic agents and the topology of road networks. The framework leverages frozen LLMs as the reasoning engine, employing a traffic encoder to extract spatial-level scene features from observed trajectories of agents, while a lightweight Convolutional Neural Network (CNN) encodes the local high-definition (HD) maps. To assess the intrinsic reasoning ability of LLMs, the extracted scene features are then transformed into LLM-compatible tokens via a reprogramming adapter. By residing the prediction burden with the LLMs, a simpler linear decoder is applied to output future trajectories. The framework enables a quantitative analysis of the influence of multi-modal information, especially the impact of map semantics on trajectory prediction accuracy, and allows seamless integration of frozen LLMs with minimal adaptation, thereby demonstrating strong generalizability across diverse LLM architectures and providing a unified platform for model evaluation.
Code and trained model checkpoints are available at \url{https://github.com/glee220/trajectory_prediction}

\end{abstract}


\section{Introduction}
Comprehending and reasoning about complex traffic scenarios is a core prerequisite for achieving safe and reliable driving maneuvers in AD research~\cite{xu2024comprehensive}. Autonomous systems are required to accurately perceive the state of surrounding traffic participants, infer their potential intentions, and make informed decisions and plans based on the road topology, all while operating in dynamic and uncertain traffic environments~\cite{teng2025improving,fu2024summary}. In real-world scenarios, the intricate interactions among traffic agents and the diverse road structures mean that the system's understanding of the scene is not only dependent on surface-level observations but also on a deeper understanding of spatio-temporal relationships~\cite{chen2024review}. Therefore, enabling models to effectively reason about these relationships and predict future trajectories is a critical scientific challenge for advancing the safety and reliability of AD systems~\cite{madjid2025trajectory}.

LLMs have recently demonstrated exceptional reasoning capabilities in the field of natural language processing~\cite{annepaka2025large}. First, self-supervised pretraining on large-scale corpora enables these models to learn rich world knowledge, semantic associations, and causal logic from vast amounts of linguistic data~\cite{qing2025semi}, thus acquiring the ability for task transfer and knowledge inference~\cite{zhang2024knowledge}. Second, the multi-head attention mechanism within the Transformer architecture allows the model to effectively capture long-range dependencies and contextual relationships~\cite{vaswani2017attention}, forming a comprehensive representation of complex event structures. Moreover, with the continuous expansion of model parameters, the generalization ability of LLMs has been significantly enhanced~\cite{berti2025emergent}, enabling them to reorganize existing knowledge and perform logical reasoning and decision-making even in previously unseen scenarios~\cite{liu2025logical, patil2025advancing}. These characteristics have enabled LLMs to exhibit human-like cognitive abilities in tasks such as language understanding, planning reasoning, and physical scene modeling~\cite{wang2023empowering}. Recent studies have explored the potential of LLMs for handling spatio-temporal data. Time-LLM \cite{jin2024timellm} reprograms time-series features into token-like embeddings so that frozen LLMs can perform forecasting without retraining. ST-LLM \cite{liu2024spatial} represents spatial and temporal information as discrete tokens and demonstrates strong few-shot performance in traffic forecasting tasks. 
Jiawei Liu et al.~\cite{liu2025harnessing} evaluated the intrinsic extrapolation ability of LLMs for vehicle trajectory prediction, but their work primarily focused on trajectory prediction itself, without considering the integration of multi-modal information, such as map data, which is crucial for understanding the spatial context and the interactions between the vehicle and its surrounding environment.

Motivated by the above considerations, this generalization mechanism based on semantics and causal reasoning provides new insights into traffic scene understanding for AD. LLMs have the potential to  represent multi-agent interactions, spatio-temporal dependencies, and road topology constraints in complex traffic environments~\cite{zhi2025lscenellm}, laying a foundation for exploring the performance and potential of LLMs in AD tasks. Therefore, before applying LLMs into safety-critical systems, it is essential to conduct a comprehensive and reliable evaluation of their spatio-temporal reasoning capabilities~\cite{chang2024survey}.

Trajectory prediction stands as an ideal benchmark for assessing these abilities. As a critical task in AD systems~\cite{bharilya2024machine}, it necessitates precise forecasting of traffic participants' future trajectories within dynamic and complex environments, enabling informed decision-making and planning~\cite{xu2025survey}. Given its reliance on spatio-temporal interactions between agents and their surroundings, trajectory prediction effectively showcases the model’s spatio-temporal reasoning proficiency.~\cite{gao2025vehicle}.

This paper proposes a framework that uses trajectory prediction as the core task for evaluating the spatio-temporal reasoning capabilities of LLMs. The framework extracts key spatio-temporal features from traffic scenes, including the movement patterns of traffic participants and road topology information, to facilitate spatio-temporal reasoning and prediction. The model's spatio-temporal reasoning ability is quantitatively evaluated through trajectory prediction error metrics, providing a quantitative assessment of its understanding and reasoning performance in complex traffic environments.

The contributions of this research are summarized as follows:
\begin{itemize}
\item A systematic evaluation framework is proposed to assess LLMs’ spatio-temporal reasoning, with a particular emphasis on integrating local HD map information alongside ego-vehicle and surrounding-agent trajectories, thereby enabling quantitative analysis of the effect of map semantics on trajectory prediction accuracy.
\item The framework enables seamless integration of frozen LLMs for trajectory prediction with minimal adaptation, demonstrating strong generalizability across various LLM architectures. Experiments with models like LLaMA2, LLaMA3, and others validate its effectiveness, providing a unified platform for model evaluation.
\end{itemize}

\begin{figure*}[h]
    \centering
    \includegraphics[width=1\linewidth, trim=0.5cm 4.5cm 2cm 3.5cm , clip]
    {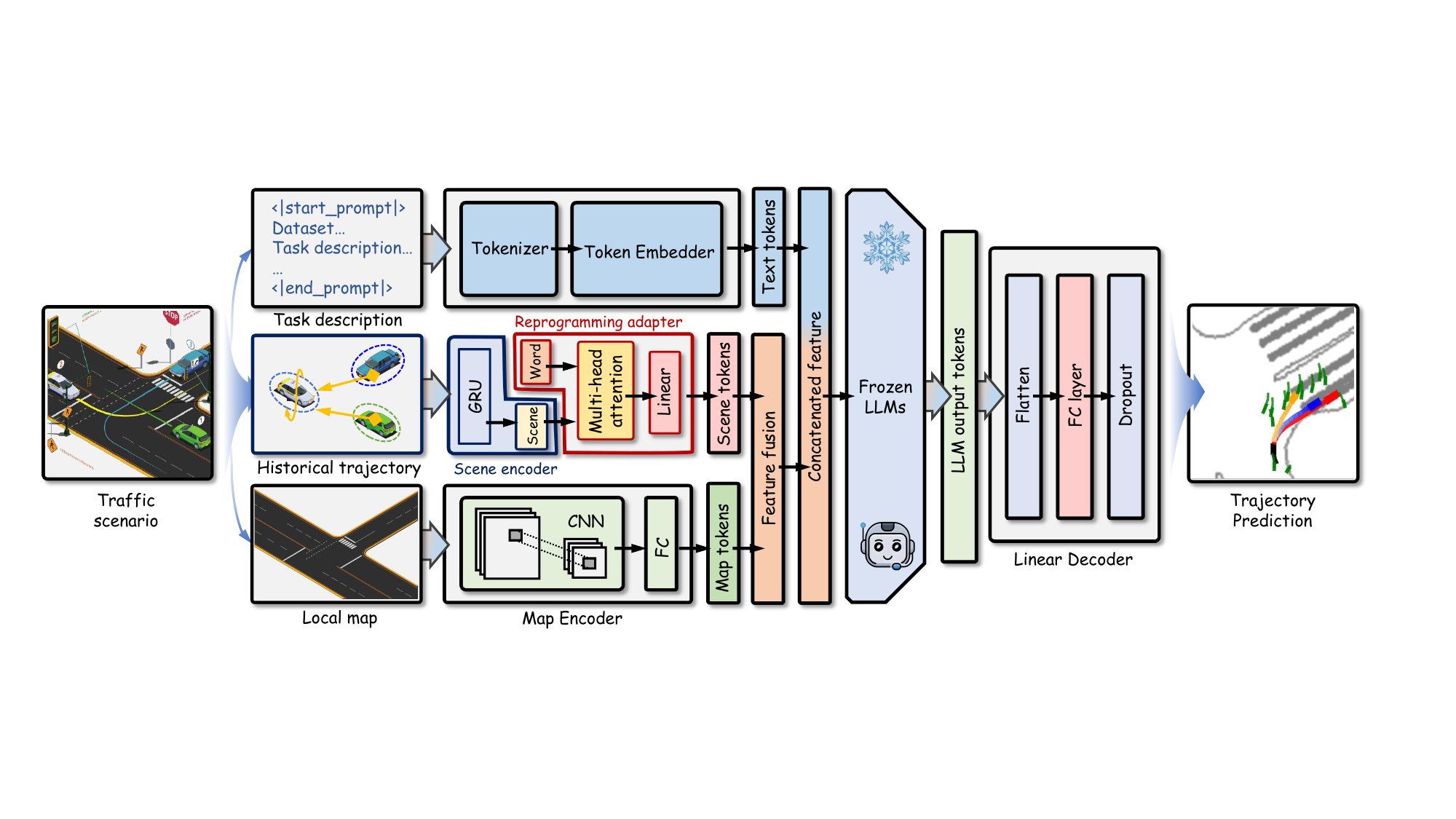}
    \caption{\centering Overview of the proposed multi-modal evaluation framework. The model integrates ego vehicle trajectories, neighboring vehicle trajectories, and local HD maps, processed by frozen LLMs to predict future trajectories.}
    \label{fig:framework}
\end{figure*}
\section{Methodology}
As the extension of our previous research \cite{liu2025harnessing}, which explored the extrapolation abilities of LLMs for vehicle trajectory prediction but did not fully integrate map data, this study addresses this gap by introducing a comprehensive framework for evaluating LLMs’ spatio-temporal reasoning, incorporating map semantics as a key component. We extend our methodology by seamlessly integrating dynamic trajectory data with static HD map semantics in a unified trajectory prediction pipeline, leveraging frozen LLMs. 

An overview of the framework is shown in
Fig.~\ref{fig:framework}.
Initially, a traffic scene encoder extracts spatial representations from the observed trajectories of traffic participants, capturing their spatial configurations and mutual interactions at each time step. Subsequently, a reprogramming adapter projects these encoded scene features into the textual embedding space, converting them into compositions of tokens from the LLM’s vocabulary. The resulting transformed features are denoted as scene tokens.
In parallel, a map encoder processes the ego vehicle's local HD map using a lightweight CNN. This module extracts lane geometry, connectivity, and topological priors, which provide static spatial context for reasoning. The resulting outputs are termed map tokens.
Next, the map tokens are fused with the previously generated scene tokens, and then concatenated with the text tokens describing the prediction task, forming a combined sequence that is fed into a frozen LLM backbone
Finally, a linear decoder, implemented as a matrix transformation, converts the output hidden states of the LLM into the predicted future trajectories.
The following subsections describe each component in detail.

\subsection{Scene Encoder:}

Efficient and precise traffic scene encoding forms the foundation for trajectory prediction, enabling the comprehensive capture of social interactions among traffic agents and the evolution of their motion relationships over time. The encoder employed in our framework is designed to extract the spatial configuration and interaction dynamics of surrounding traffic entities from time-series trajectories.

Specifically, the Encoder first vectorizes the spatial states of the ego vehicle \(0\) and its surrounding participants \( i \in \{1, 2, ..., I\} \) at each timestamp \( t \in \{1, 2, ..., T\} \) as:
\begin{equation}
\vec{x}_t^i = [x_t^i - x_{t-1}^i; \, x_t^i - x_t^0],
\end{equation}
where \( x_t^i \in \mathbb{R}^2 \) denotes the spatial coordinate, and \( I \) represents the number of surrounding participants, and \( T \) represents the number of observed timestamps.  
The resulting vectorized state \( \vec{x}_t^i \in \mathbb{R}^4 \) integrates both the displacement vector of agent \( i \) and its relative position to the ego vehicle. To ensure translation and rotation invariance, all state vectors are rotated to align with the orientation of the target vehicle.

Next, a cross-attention layer is employed to encode the interactions between the ego vehicle and its neighbors into the representation \( h_t' \in \mathbb{R}^{d_{scene}} \), defined as:
\begin{equation}
h_t' = \sum_{i=1}^I \text{Softmax}\left(\frac{q_t^0 k_t^i}{\sqrt{d_{scene}}}\right) v_t^i, \tag{2}
\end{equation}
The query vector \( q_t^0 \in \mathbb{R}^{d_{scene}} \) is derived from the target vehicle \(0\), while the key vectors \( k_t^i \in \mathbb{R}^{d_{scene}} \) and value vectors \( v_t^i \in \mathbb{R}^{d_{scene}} \) are computed from each neighboring agent \( i \) as follows:
\begin{equation}
q_t^0 = W_Q \Phi(\vec{x}_t^0), \quad 
k_t^i = W_K \Phi(\vec{x}_t^i), \quad 
v_t^i = W_V \Phi(\vec{x}_t^i), \tag{3}
\end{equation}
where \( \Phi : \mathbb{R}^4 \rightarrow \mathbb{R}^{d_{scene}} \) is a multi-layer fully connected network (FCN), and \( W_Q, W_K, W_V \in \mathbb{R}^{d_{scene} \times d_{scene}} \) are learnable projection matrices for queries, keys, and values, respectively.

Finally, a learnable fusion scheme integrates the encoded interactions of neighboring agents with the target vehicle’s own spatial state as:
\begin{equation}
h_t = \text{Sigmoid}(\alpha) \circ h_t' + \text{Sigmoid}(1 - \alpha) \circ h_t^0, \tag{4}
\label{eq:feature_fusion}
\end{equation}
where \( h_t^0 = \Phi(\vec{x}_t^0) \), \( \alpha \in \mathbb{R}^{d_{scene}} \) is a learnable gating vector, and \(\circ\) denotes the Hadamard product.

\subsection{Reprogramming Adapter:}

The purpose of the Reprogramming Adapter is to bridge the gap between raw scene-level spatio-temporal features and the LLM’s ability to process and reason over textual data. It maps the extracted scene feature \( h_t \) from Equation~\eqref{eq:feature_fusion} into the LLM's textual embedding space. Expressed as:

\[
b^{\text{scene}}_t = \text{Reprogram}(h_t), \tag{5}
\]

The mapping function is defined as \( \text{Reprogram}: \mathbb{R}^{d_{\text{scene}}} \rightarrow \mathbb{R}^{d_{\text{llm}}} \), transforming the spatio-temporal feature into a set of tokens compatible with the LLM’s input representation, represented by \( b^{\text{scene}}_t \in \mathbb{R}^{d_{\text{llm}}} \). This enables the LLM to process and reason about dynamic traffic scenes and the temporal evolution of agent interactions.

\subsection{Map Encoder:}

In AD, static road topology significantly influences vehicle motion and trajectory feasibility. To enable LLMs to effectively incorporate these spatial constraints, we introduce a lightweight Map Encoder to extract semantic features from local HD maps.

For each scene, a local map patch centered around the ego vehicle is cropped from the global HD map, ensuring spatial consistency with the vehicle's perception range. This patch includes key geometric features, such as lane dividers, and intersections. A lightweight CNN is then used to extract hierarchical spatial features from this region, producing a compact representation of the local road topology:

\[
h^{\text{map}} = f_{\text{CNN}}(M_{\text{local}}), \quad h^{\text{map}} \in \mathbb{R}^{d_{\text{map}}},
\tag{6}
\]
where \( M_{\text{local}} \) represents the rasterized local HD map, and \( f_{\text{CNN}}(\cdot) \) denotes the convolutional feature extractor.

\subsection{Feature Fusion Module:}

In the Feature Fusion Module, the goal is to combine the temporal trajectory features and spatial map features into a comprehensive scene representation for reasoning and prediction.
The cross-attention mechanism is applied between the trajectory features \( h_t^{\text{traj}} \) (denoted as \( \text{reprogrammed\_rep} \)) and the static map features \( h^{\text{map}} \). This allows the model to learn the relevant spatial correlations between the vehicle's motion and surrounding road geometry, focusing on key areas such as lane exits, curves, and intersections. The output of the cross-attention operation is computed as:

\[
\tilde{h}_t^{\text{map}} = \text{MHA}(h_t^{\text{traj}}, h^{\text{map}}, h^{\text{map}}),
\tag{7}
\]

After the attention mechanism, the map-related features \( \tilde{h}_t^{\text{map}} \) and the original trajectory features \( h_t^{\text{traj}} \) are concatenated along the feature dimension. The concatenated features are then passed through a linear layer for fusion. This fusion produces a comprehensive scene representation \( f_t \), which integrates both the spatial topology of the road and the dynamic interactions of the agents:

\[
f_t = \text{Linear}(\text{concat}(\tilde{h}_t^{\text{map}}, h_t^{\text{traj}})),
\tag{8}
\]

The fused feature \( f_t \) is then combined with the prompt embeddings that describe the task, creating the final input for the LLM. This ensures that the model has access to both the spatial and dynamic features for subsequent reasoning.

\subsection{Linear Decoder:}  
Employing autoregressive architectures to decode LLM outputs may inadvertently shift the prediction complexity to the decoder itself, thereby obscuring the evaluation of the LLM’s inherent reasoning capability. To prevent delegating the prediction task to a sophisticated decoder, the proposed framework adopts a lightweight linear model to decode the LLM’s outputs. This approach is formulated as:

\[
\hat{\tau}_{1:N}^{0} = \text{LinearDecoder}(\{e_{t}^{\text{scene}}\}_{t=1}^{T}),
\tag{9}
\]

Here, \( e_{t}^{\text{scene}} \in \mathbb{R}^{d_{\text{llm}}} \) represents the scene token \( b_{t}^{\text{scene}} \) after being processed by the frozen LLM backbone. The Linear Decoder flattens the sequence \( \{e_{t}^{\text{scene}}\}_{t=1}^{T} \) into a single vector and then projects it onto the predicted trajectory \( \hat{\tau}_{1:N}^{0} \in \mathbb{R}^{N \times 2} \) using a learnable weight matrix \( W_{\text{decode}} \in \mathbb{R}^{T d_{\text{llm}} \times 2N} \), where \( N \) represents the number of prediction timestamps.

This lightweight decoding strategy ensures that the prediction complexity remains manageable, thus allowing a more direct assessment of the LLM’s intrinsic reasoning abilities.

\section{Experiments}

To validate the effectiveness of the proposed framework, ablation experiments were conducted to analyze the model's reasoning capabilities. The complexity of trajectory prediction was increased step by step, starting with the target vehicle’s trajectory, then adding neighboring vehicle trajectories, and finally incorporating local map information. These experiments assessed how the model integrates multi-modal information to optimize prediction performance. Additionally, the role of local map information in improving prediction accuracy was thoroughly analyzed. To further assess the framework's generalizability, we conducted comparative experiments with several widely-used LLMs, evaluating their performance in trajectory prediction.

\subsection{Experimental Setup}
Experiments are conducted on nuScenes dataset~\cite{caesar2020nuscenes}, which provides annotated trajectories, surrounding agents, and HD maps. The trajectories are sampled at 2 Hz and used to predict the future 6 seconds based on the previous 2 seconds.

Model performance is evaluated using Average Displacement Error (ADE), Final Displacement Error (FDE), Missing Rate (MR), and Inference Efficiency (IE). ADE and FDE measure prediction accuracy, while MR evaluates the proportion of predicted positions deviating by more than 2 meters from the ground truth. IE assesses the model's inference efficiency, which is crucial for real-time AD applications.

\subsection{Ablation Study on multi-modal Fusion}

\begin{figure*}[htbp]
    \centering
    \includegraphics[width=0.98\textwidth]{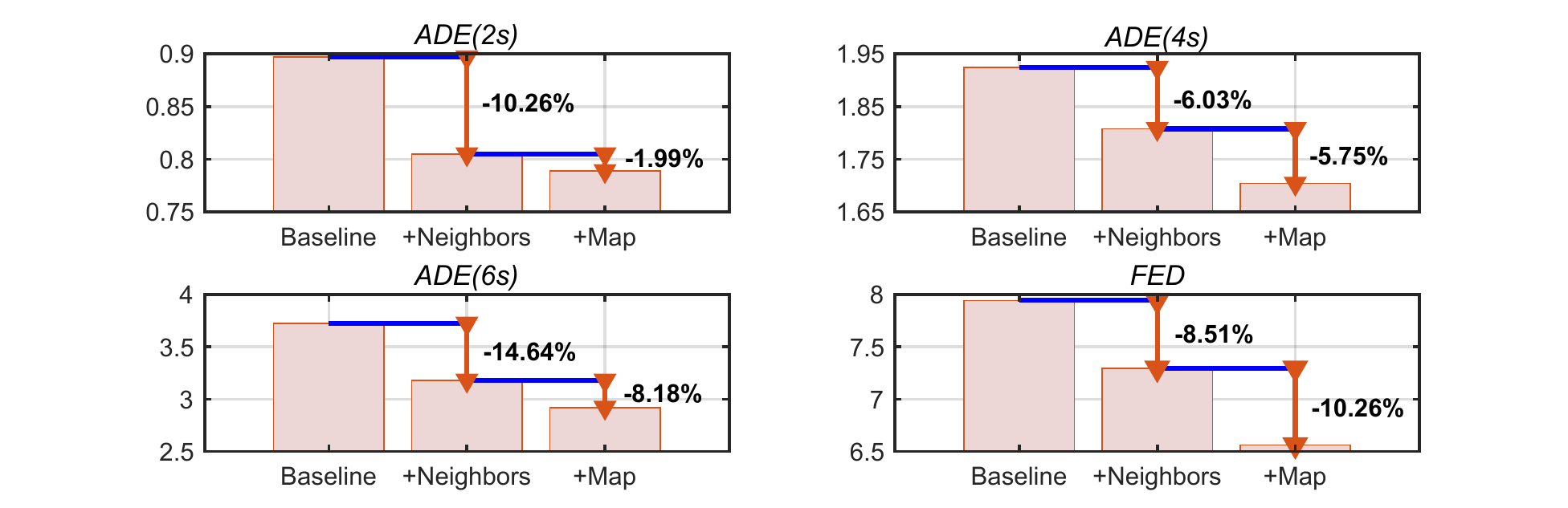}
    \caption{\centering Ablation Comparison of ADE and FDE at different time horizons.}
    \label{fig:ablation_figure}
\end{figure*}

\begin{figure*}[htbp]
    \centering
    \includegraphics[width=0.95\textwidth]{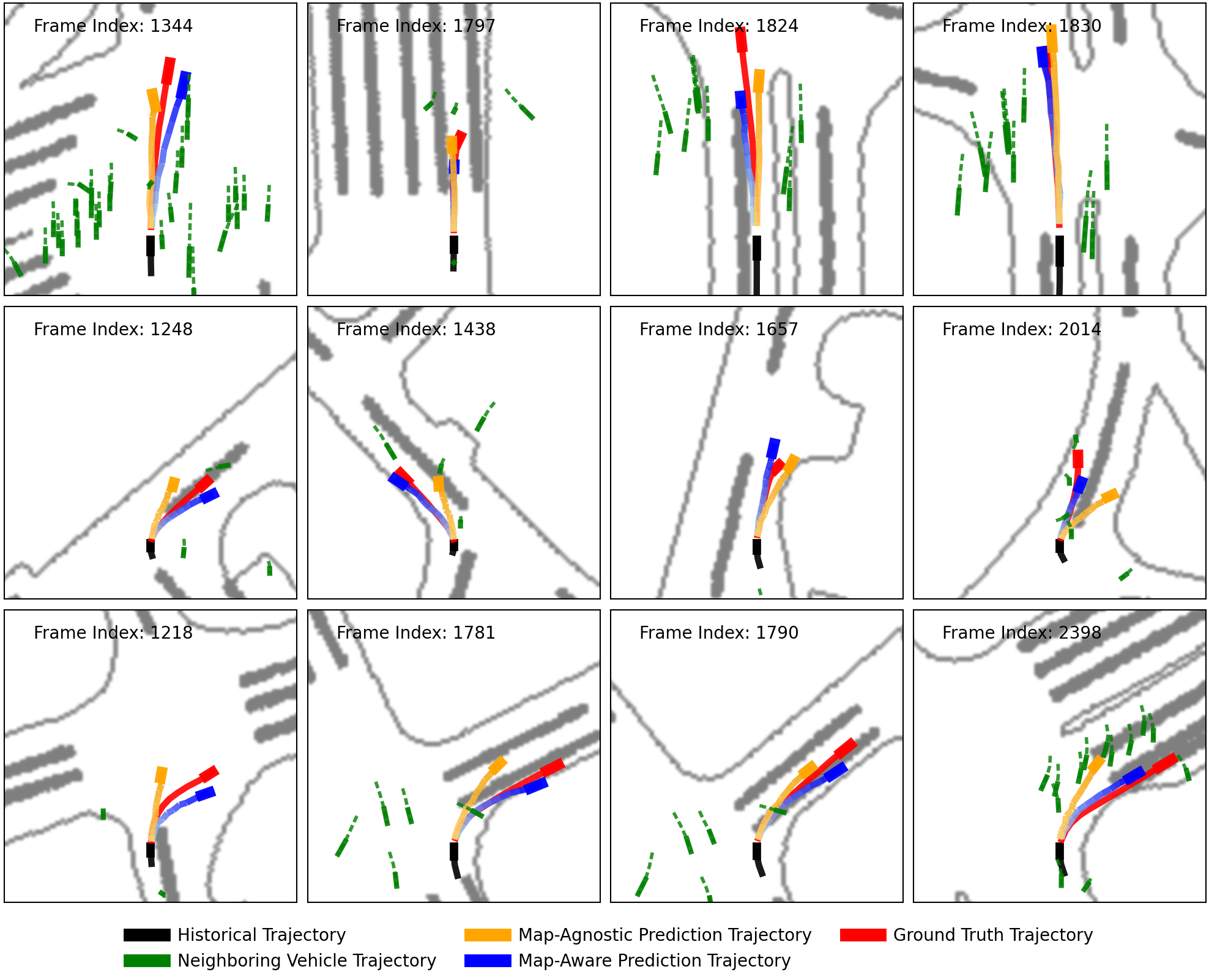}
    \caption{\centering Visualization of map-aware trajectory predictions.
    The first row represents straight scenarios, \\ the second row represents turning scenarios, and the third row represents intersection scenarios.}
    \label{fig:vis}
\end{figure*}

To rigorously evaluate the impact of each information modality, we conduct an ablation study with three configurations: 'Ego only,' using the ego vehicle's trajectory; 'Ego + Neighbor,' incorporating neighboring vehicles’ trajectories; and 'Ego + Neighbor + Map,' which also includes local HD map information. The results, shown in Table~\ref{tab:table1} and Fig.~\ref{fig:ablation_figure}, reveal that adding neighboring vehicle trajectories reduces both ADE and FDE across all time horizons, with 'Ego + Neighbor' yielding the second-best performance, as marked in blue. Incorporating map information further improves accuracy, achieving the lowest ADE and FDE values, highlighting the importance of road topology in trajectory prediction, as marked in red. However, this improvement comes with a slight increase in IE, from 0.019 seconds for 'Ego only' to 0.037 seconds for 'Ego + Neighbor + Map.' Fig.~\ref{fig:vis} demonstrates the trajectory prediction results, emphasizing the contribution of map understanding in various scenarios.

\begin{figure*}[htbp]
    \centering
    \includegraphics[width=0.98\textwidth]{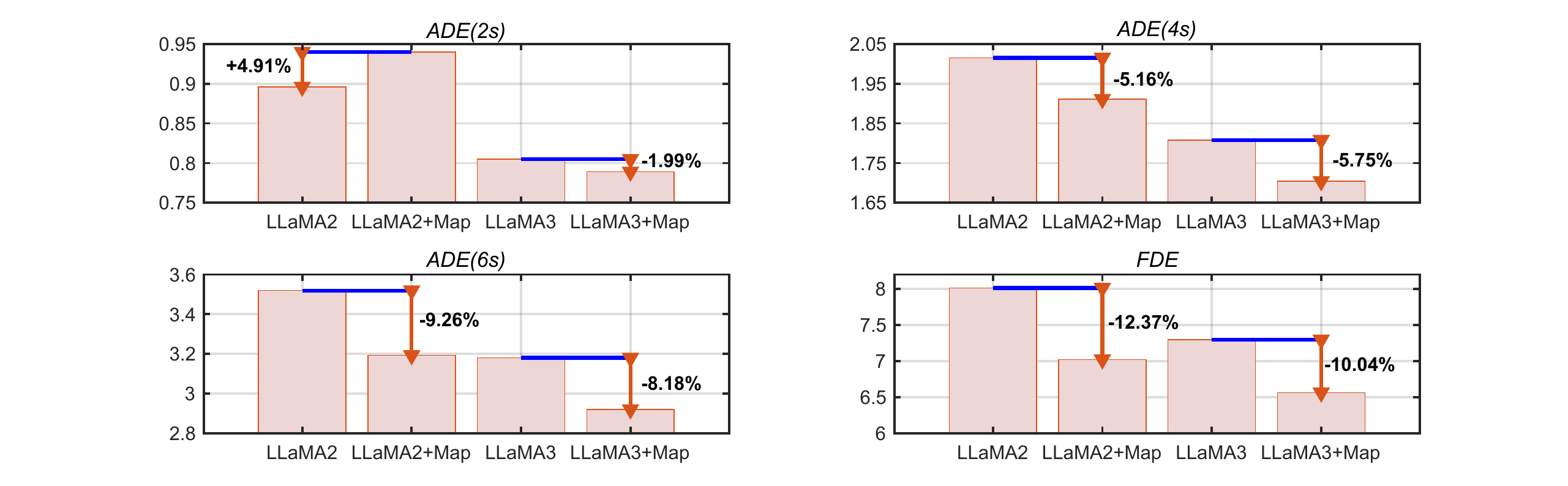}
    \caption{\centering Comparison of LLaMA2 and LLaMA3 with/without utilizing Map for ADE and FDE across Different Time Horizons.}
    \label{fig:ablation_map_figures}
\end{figure*}

Through further investigation, we have quantitatively analyzed the improvement in prediction accuracy when local map information is incorporated into the model. Show in Table~\ref{tab:table2} and Fig.~\ref{fig:ablation_map_figures}. Based on the Ego + Neighbor configuration, we examine the performance of LLaMA2 and LLaMA3 with and without integrating map information. In the 2s ADE, incorporating map information did not result in a significant reduction in error, and even led to a 4.91\% increase in the ADE for LLaMA2. However, as the prediction horizon extended, the map's benefit became apparent, improving both ADE and FDE at the 4s and 6s marks. Over time, This indicates that LLMs gradually captured the road topology's constraints, leading to more accurate predictions.
\begin{figure}[H]
    \centering
    \includegraphics[width=0.48\textwidth]{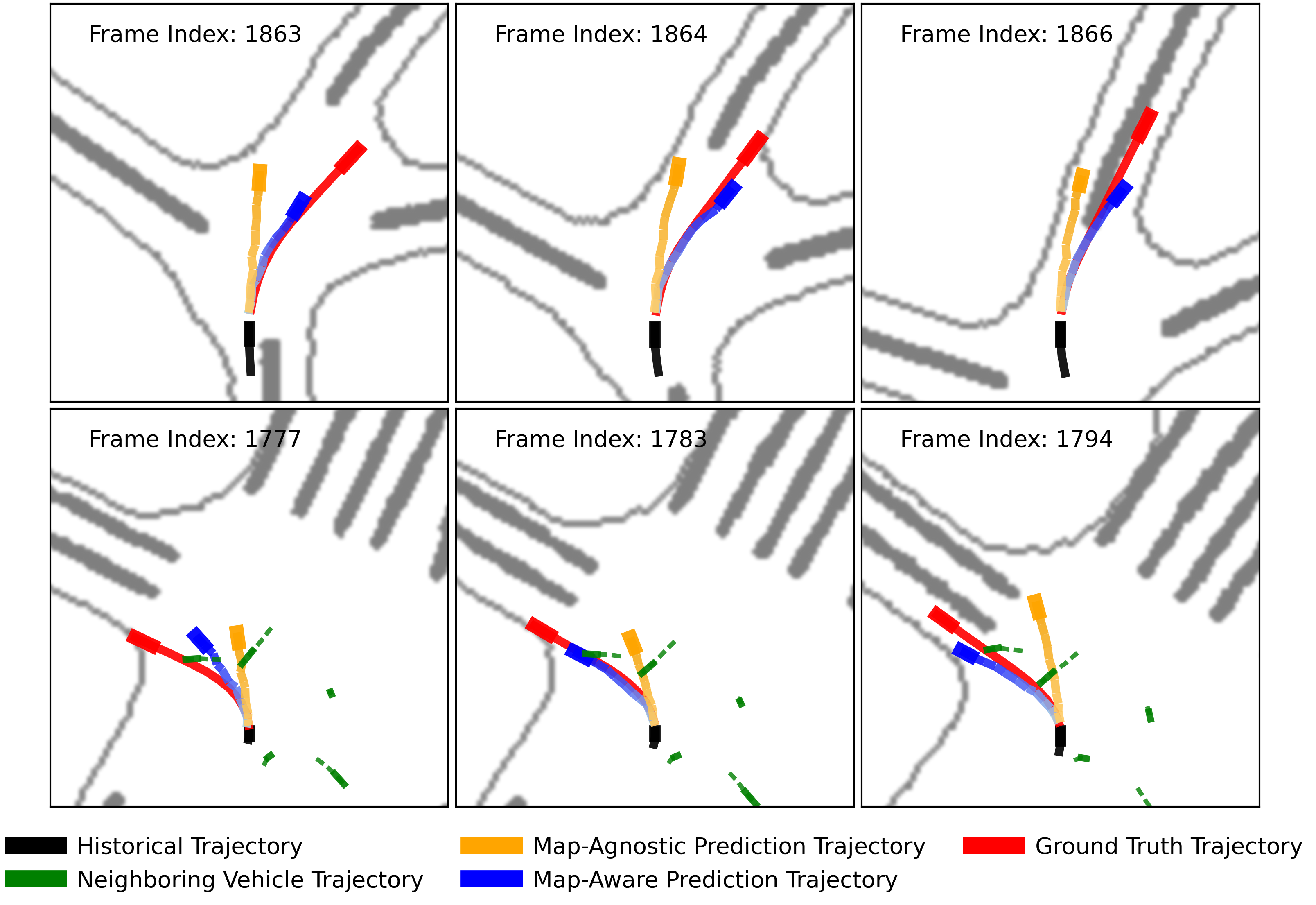}
    \caption{\centering Turning Scenarios.}
    \label{fig:turning_scenarios}
\end{figure}
As shown in Fig.~\ref{fig:turning_scenarios}, in turning scenarios, LLMs base their predictions on the historical trajectories from the outset to estimate the target lane orientation. The constraints from the map are not yet manifested. As the vehicle moves, the model adjusts its trajectory in accordance with the road structure, eventually entering the correct lane. The influence of the map becomes more evident in these scenarios, as shown by the fact that the blue one, which incorporates map information, is closer to the red ground-truth trajectory compared to the orange trajectory, which does not utilize map information.

\begin{table*}[htbp]
\caption{Ablation study on the impact of input modalities on trajectory prediction performance.}
\begin{center}
\begin{tabular}{|c|c|c|c|c|c|}
\hline
\multirow{2}{*}{\textbf{Modality}} &\multicolumn{5}{|c|}{\textbf{Evaluation Metrics}} \\
\cline{2-6} 
& \textbf{\textit{ADE±STD(2s)}} & \textbf{\textit{ADE±STD(4s)}} & \textbf{\textit{ADE±STD(6s)}} & \textbf{\textit{FDE±STD(6s)}} & \textbf{\textit{IE (s)}} \\
\hline
Ego only & 0.897 $\pm$ 1.123 & 1.924 $\pm$ 2.533 & 3.725 $\pm$ 4.983 & 7.945 $\pm$ 8.531 & \textcolor{red} {0.019} \\
\hline
Ego + Neighbor & \textcolor{blue}{0.805 $\pm$ 1.058} & \textcolor{blue}{1.808 $\pm$ 2.490} & \textcolor{blue}{3.180 $\pm$ 4.460} & \textcolor{blue}{7.296 $\pm$ 7.143 } & \textcolor{blue}{0.020} \\
\hline
Ego + Neighbor + Map & \textcolor{red}{0.789 $\pm$ 1.038 }& \textcolor{red}{1.704 $\pm$ 2.385} & \textcolor{red}{2.920 $\pm$ 4.202} & \textcolor{red}{6.563 $\pm$ 6.802} & {0.037} \\
\hline
\end{tabular}
\label{tab:table1}
\end{center}
\end{table*}

\begin{table*}[htbp]
\caption{Ablation study on map utilization.}
\begin{center}
\begin{tabular}{|c|c|c|c|c|c|c|}
\hline
\multirow{2}{*}{\textbf{Model}}&\multicolumn{6}{|c|}{\textbf{Evaluation Metrics}} \\
\cline{2-7} 
& \textbf{\textit{ADE±STD(2s)}}& \textbf{\textit{ADE±STD(4s)}}& \textbf{\textit{ADE±STD(6s)}}& \textbf{\textit{FDE±STD(6s)}}& \textbf{\textit{MR (\%)}}& \textbf{\textit{IE (s)}} \\
\hline
LLaMA2 & 0.896 $\pm$ 1.169 & 2.015 $\pm$ 2.746 & 3.519 $\pm$ 4.906 & 8.013 $\pm$ 7.864 & 67.77 & 0.035 \\
\hline
\rowcolor{gray!20} LLaMA2 + Map & 0.940 $\pm$ 1.169 & 1.911 $\pm$ 2.581 & 3.193 $\pm$ 4.455 & 7.022 $\pm$ 7.127 & 66.36 & 0.036 \\
\hline
LLaMA3 & 0.805 $\pm$ 1.058 & 1.808 $\pm$ 2.490 & 3.180 $\pm$ 4.460 & 7.296 $\pm$ 7.143 & 67.20 & 0.036 \\
\hline
\rowcolor{gray!20} LLaMA3 + Map & 0.789 $\pm$ 1.038 & 1.704 $\pm$ 2.385 & 2.920 $\pm$ 4.202 & 6.563 $\pm$ 6.802 & 65.35 & 0.037 \\
\hline
\end{tabular}
\label{tab:table2}
\end{center}
\end{table*}

\begin{figure*}[htbp]
    \centering
    \includegraphics[width=0.95\textwidth]{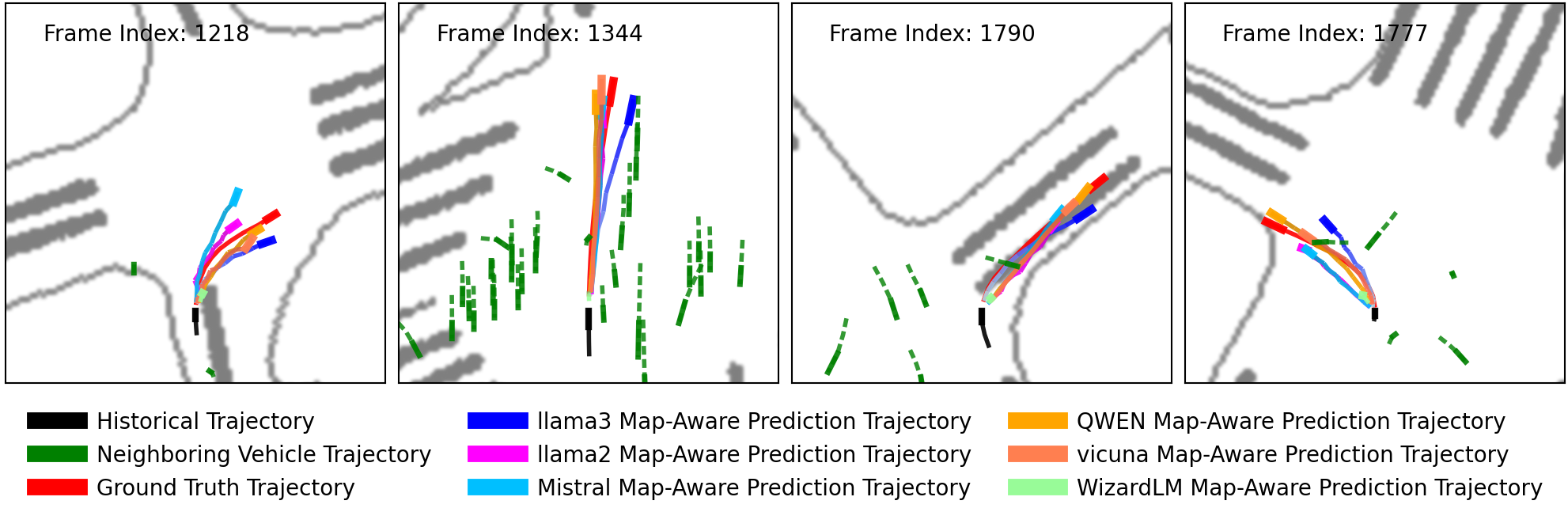}
    \caption{\centering Generalizability evaluation across six LLM backbones.}
    \label{fig:cross-model}
\end{figure*}


\begin{table*}[htbp]
\caption{generalizability evaluation across six LLM backbones.(Ego + Neighbor + Map).}
\begin{center}
\begin{tabular}{|c|c|c|c|c|c|c|}
\hline
\multirow{2}{*}{\textbf{Model}}&\multicolumn{6}{|c|}{\textbf{Evaluation Metrics}} \\
\cline{2-7} 
& \textbf{\textit{ADE±STD(2s)}}& \textbf{\textit{ADE±STD(4s)}}& \textbf{\textit{ADE±STD(6s)}}& \textbf{\textit{FDE±STD(6s)}}& \textbf{\textit{MR (\%)}}& \textbf{\textit{IE(s)}} \\
\hline
LLaMA2 & 0.940 $\pm$ 1.169 & 1.911 $\pm$ 2.581 & 3.193 $\pm$ 4.455 & 7.022 $\pm$ 7.127 & 66.36 & 0.036 \\
\hline
LLaMA3 & 0.789 $\pm$ 1.038 & 1.704 $\pm$ 2.385 & 2.920 $\pm$ 4.202 & 6.563 $\pm$ 6.802 & 65.35 & 0.037 \\
\hline
Qwen2.5 & 0.822 $\pm$ 1.071 & 1.762 $\pm$ 2.445 & 3.002 $\pm$ 4.260 & 6.710 $\pm$ 6.826 & 65.97 & 0.034 \\
\hline
Mistral & 0.825 $\pm$ 1.069 & 1.766 $\pm$ 2.441 & 3.001 $\pm$ 4.270 & 6.686 $\pm$ 6.878 & 65.84 & 0.035 \\
\hline
Vicuna & 0.816 $\pm$ 1.049 & 1.730 $\pm$ 2.389 & 2.938 $\pm$ 4.184 & 6.551 $\pm$ 6.751 & 64.81 & 0.035 \\
\hline
WizardLM & 3.151 $\pm$ 4.214 & 6.667 $\pm$ 9.149 & 10.404 $\pm$ 14.092 & 20.816 $\pm$ 21.110 & 68.37 & 0.035 \\
\hline
\end{tabular}
\label{tab:table3}
\end{center}
\end{table*}

\subsection{Evaluating the Generalizability of Our Model Across Different LLMs in Trajectory Prediction}

One of the core advantages of the proposed framework is its minimal adaptation requirement. By simply adjusting the dimensionality of the LLM input parameters and configuration settings, off-the-shelf LLMs can be seamlessly integrated into the trajectory prediction task with almost no additional modifications. This design endows the framework with strong generalizability, enabling it to easily accommodate various LLM architectures. 
To evaluate the generalizability of the proposed framework across diverse LLMs. This paper adopts several commonly used LLMs, including LLaMA2, LLaMA3, Qwen2.5, Mistral, Vicuna, and WizardLM, to test the generalizability of the proposed framework. The results are shown in Table~\ref{tab:table3} and Fig.~\ref{fig:cross-model}.
Experimental results demonstrate the framework's performance across multiple LLMs. The framework effectively distinguishes the performance of different models, showing that it is capable of handling varying model architectures and evaluating them on a unified platform.

These results validate the generalizability of our framework. It provides a unified standard for evaluating different LLMs and offers critical insights into model selection, and with the continued advancement of LLMs, the framework holds the potential to leverage these models for even better performance in the future.


\section{Conclusion}
This study introduces a framework for assessing the spatio-temporal reasoning abilities of LLMs, with a particular focus on trajectory prediction tasks in AD. By incorporating trajectories of dynamic traffic agents and static HD map semantics, experimental findings illustrate that the framework proficiently integrates multi-modal data, with a particular emphasis on analyzing the impact of map semantics.

Through experiments with various LLMs, the results validate that the framework demonstrates robust generalizability across different models, requiring minimal adaptation. Consequently, the proposed framework not only serves as a powerful tool for evaluating the spatio-temporal reasoning capacities of LLMs in complex traffic scenarios but also has the potential to harness increasingly advanced reasoning abilities as LLMs evolve. This capability could play a pivotal role in advancing AD systems, propelling intelligent transportation systems toward higher levels of intelligence and safety.

\bibliographystyle{IEEEtran}
\bibliography{references}

@article{teng2025improving,
  title={Improving intelligent perception and decision optimization of pedestrian crossing scenarios in autonomous driving environments through large visual language models},
  author={Teng, Xiao and Huang, Lin and Shen, Zhenjiang and Li, Wankai},
  journal={Scientific Reports},
  volume={15},
  number={1},
  pages={31283},
  year={2025},
  publisher={Nature Publishing Group UK London}
}

@article{xu2024comprehensive,
  title={A comprehensive review of autonomous driving algorithms: Tackling adverse weather conditions, unpredictable traffic violations, blind spot monitoring, and emergency maneuvers},
  author={Xu, Cong and Sankar, Ravi},
  journal={Algorithms},
  volume={17},
  number={11},
  pages={526},
  year={2024},
  publisher={MDPI}
}

@article{xu2025survey,
  title={A Survey of Autonomous Driving Trajectory Prediction: Methodologies, Challenges, and Future Prospects.},
  author={Xu, Miao and Liu, Zhi and Wang, Bingyi and Li, Shengyan},
  journal={Machines},
  volume={13},
  number={9},
  year={2025}
}

@article{chen2024review,
  title={A review of decision-making and planning for autonomous vehicles in intersection environments},
  author={Chen, Shanzhi and Hu, Xinghua and Zhao, Jiahao and Wang, Ran and Qiao, Min},
  journal={World Electric Vehicle Journal},
  volume={15},
  number={3},
  pages={99},
  year={2024},
  publisher={MDPI}
}

@article{madjid2025trajectory,
  title={Trajectory prediction for autonomous driving: Progress, limitations, and future directions},
  author={Madjid, Nadya Abdel and Ahmad, Abdulrahman and Mebrahtu, Murad and Babaa, Yousef and Nasser, Abdelmoamen and Malik, Sumbal and Hassan, Bilal and Werghi, Naoufel and Dias, Jorge and Khonji, Majid},
  journal={arXiv preprint arXiv:2503.03262},
  year={2025}
}

@article{bharilya2024machine,
  title={Machine learning for autonomous vehicle's trajectory prediction: A comprehensive survey, challenges, and future research directions},
  author={Bharilya, Vibha and Kumar, Neetesh},
  journal={Vehicular Communications},
  volume={46},
  pages={100733},
  year={2024},
  publisher={Elsevier}
}

@article{annepaka2025large,
  title={Large language models: a survey of their development, capabilities, and applications},
  author={Annepaka, Yadagiri and Pakray, Partha},
  journal={Knowledge and Information Systems},
  volume={67},
  number={3},
  pages={2967--3022},
  year={2025},
  publisher={Springer}
}

@article{qing2025semi,
  title={Semi-supervised feature selection with minimal redundancy based on group optimization strategy for multi-label data},
  author={Qing, Depeng and Zheng, Yifeng and Zhang, Wenjie and Ren, Weishuo and Zeng, Xianlong and Li, Guohe},
  journal={Knowledge and Information Systems},
  volume={67},
  number={2},
  pages={1271--1308},
  year={2025},
  publisher={Springer}
}

@article{vaswani2017attention,
  title={Attention is all you need},
  author={Vaswani, Ashish and Shazeer, Noam and Parmar, Niki and Uszkoreit, Jakob and Jones, Llion and Gomez, Aidan N and Kaiser, {\L}ukasz and Polosukhin, Illia},
  journal={Advances in neural information processing systems},
  volume={30},
  year={2017}
}

@inproceedings{jin2024timellm,
  title     = {Time-LLM: Time Series Forecasting by Reprogramming Large Language Models},
  author    = {Jin, Ming and Wang, Shiyu and Ma, Lintao and Chu, Zhixuan and Zhang, James Y. and Shi, Xiaoming and Chen, Pin-Yu and Liang, Yuxuan and Li, Yuan-Fang and Pan, Shirui and Wen, Qingsong},
  booktitle = {International Conference on Learning Representations (ICLR)},
  year      = {2024},
  url       = {https://arxiv.org/abs/2310.01728},
  note      = {Accepted at ICLR 2024}
}

@inproceedings{liu2024spatial,
  title={Spatial-temporal large language model for traffic prediction},
  author={Liu, Chenxi and Yang, Sun and Xu, Qianxiong and Li, Zhishuai and Long, Cheng and Li, Ziyue and Zhao, Rui},
  booktitle={2024 25th IEEE International Conference on Mobile Data Management (MDM)},
  pages={31--40},
  year={2024},
  organization={IEEE}
}

@article{liu2025logical,
  title={Logical reasoning in large language models: A survey},
  author={Liu, Hanmeng and Fu, Zhizhang and Ding, Mengru and Ning, Ruoxi and Zhang, Chaoli and Liu, Xiaozhang and Zhang, Yue},
  journal={arXiv preprint arXiv:2502.09100},
  year={2025}
}

@article{patil2025advancing,
  title={Advancing reasoning in large language models: Promising methods and approaches},
  author={Patil, Avinash and Jadon, Aryan},
  journal={arXiv preprint arXiv:2502.03671},
  year={2025}
}

@inproceedings{zhi2025lscenellm,
  title={Lscenellm: Enhancing large 3d scene understanding using adaptive visual preferences},
  author={Zhi, Hongyan and Chen, Peihao and Li, Junyan and Ma, Shuailei and Sun, Xinyu and Xiang, Tianhang and Lei, Yinjie and Tan, Mingkui and Gan, Chuang},
  booktitle={Proceedings of the Computer Vision and Pattern Recognition Conference},
  pages={3761--3771},
  year={2025}
}

@article{fu2024summary,
  title={Summary and reflections on pedestrian trajectory prediction in the field of autonomous driving},
  author={Fu, Zheng and Jiang, Kun and Xie, Chuchu and Xu, Yuhang and Huang, Jin and Yang, Diange},
  journal={IEEE Transactions on Intelligent Vehicles},
  year={2024},
  publisher={IEEE}
}

@article{gao2025vehicle,
  title={A vehicle trajectory prediction model that integrates spatial interaction and multiscale temporal features},
  author={Gao, Yuan and Yang, Kaifeng and Yue, Yibing and Wu, Yunfeng},
  journal={Scientific Reports},
  volume={15},
  number={1},
  pages={8217},
  year={2025},
  publisher={Nature Publishing Group UK London}
}

@inproceedings{liu2025harnessing,
  title     = {Harnessing and Evaluating the Intrinsic Extrapolation Ability of Large Language Models for Vehicle Trajectory Prediction},
  author    = {Liu, Jie and Liu, Yang and Gong, Xinyu and Wang, Tianyu and Chen, Hui and Hu, Yeping},
  booktitle = {Proceedings of the 2025 Conference of the North American Chapter of the Association for Computational Linguistics: Human Language Technologies},
  pages     = {4379--4391},
  year      = {2025}
}

@inproceedings{caesar2020nuscenes,
  title={nuscenes: A multimodal dataset for autonomous driving},
  author={Caesar, Holger and Bankiti, Varun and Lang, Alex H and Vora, Sourabh and Liong, Venice Erin and Xu, Qiang and Krishnan, Anush and Pan, Yu and Baldan, Giancarlo and Beijbom, Oscar},
  booktitle={Proceedings of the IEEE/CVF conference on computer vision and pattern recognition},
  pages={11621--11631},
  year={2020}
}

@inproceedings{zhang2024knowledge,
  title={Knowledge and Task-Driven Multimodal Adaptive Transfer Through LLMs with Limited Data},
  author={Zhang, Xu and Chen, Zhikui and Ren, Hao and Tian, Yang},
  booktitle={2024 IEEE International Conference on Bioinformatics and Biomedicine (BIBM)},
  pages={5343--5348},
  year={2024},
  organization={IEEE}
}

@article{berti2025emergent,
  title={Emergent abilities in large language models: A survey},
  author={Berti, Leonardo and Giorgi, Flavio and Kasneci, Gjergji},
  journal={arXiv preprint arXiv:2503.05788},
  year={2025}
}

@article{wang2023empowering,
  title={Empowering autonomous driving with large language models: A safety perspective},
  author={Wang, Yixuan and Jiao, Ruochen and Zhan, Sinong Simon and Lang, Chengtian and Huang, Chao and Wang, Zhaoran and Yang, Zhuoran and Zhu, Qi},
  journal={arXiv preprint arXiv:2312.00812},
  year={2023}
}

@article{chang2024survey,
  title={A survey on evaluation of large language models},
  author={Chang, Yupeng and Wang, Xu and Wang, Jindong and Wu, Yuan and Yang, Linyi and Zhu, Kaijie and Chen, Hao and Yi, Xiaoyuan and Wang, Cunxiang and Wang, Yidong and others},
  journal={ACM transactions on intelligent systems and technology},
  volume={15},
  number={3},
  pages={1--45},
  year={2024},
  publisher={ACM New York, NY}
}
\end{document}